\documentclass[fleqn,10pt]{wlscirep}
\usepackage[utf8]{inputenc}
\usepackage[T1]{fontenc}
\usepackage{graphicx}%
\usepackage{booktabs}
\usepackage{multirow}
\usepackage{algorithm}
\usepackage{algorithmic}
\usepackage{setspace}

\begin{document}
\doublespacing

\title{When OpenClaw Meets Hospital: Toward an Agentic Operating System for Dynamic Clinical Workflows}

\author[1]{Wenxian Yang} 
\author[2]{Hanzheng Qiu}
\author[3]{Bangqun Zhang} 
\author[4]{Chengquan Li}
\author[5]{Zhiyong Huang}
\author[6,*]{Xiaobin Feng}
\author[4,*]{Rongshan Yu} 
\author[7,*]{Jiahong Dong}

\affil[1]{Independent Researcher}
\affil[2]{National Institute for Data Science in Health and Medicine, Xiamen University, Xiamen, China}
\affil[3]{Yue'erwan Internet Hospital Co., Ltd.}
\affil[4]{School of Biomedical Engineering, Tsinghua University, Beijing, China}
\affil[5]{National University of Singapore, Singapore}
\affil[6]{Yttrium-90 Precision Interventional Radiotherapy Center of Liver Cancer, Beijing Tsinghua Changgung Hospital, School of Clinical Medicine, Tsinghua University, Beijing, China}
\affil[7]{Hepatobiliary and Pancreatic Centre, Beijing Tsinghua Changgung Hospital, School of Clinical Medicine, Tsinghua University, Beijing, China}

\affil[*]{e-mail: fengxiaobin@mail.tsinghua.edu.cn, rsyu@xmu.edu.cn, dongjiahong@mail.tsinghua.edu.cn}

\begin{abstract}
Large language model (LLM) agents extend generative models with reasoning, tool use, and persistent memory, thereby enabling the automation of complex tasks. In healthcare, such systems could support documentation, care coordination, and clinical decision making. Their reliable deployment in hospitals, however, remains constrained by safety risks, limited transparency, and inadequate mechanisms for handling longitudinal clinical context. Here we propose an architecture that adapts LLM agents to hospital environments. The design comprises four components: a restricted execution environment inspired by multi-user operating systems, a document-centric interaction model linking patient and clinician agents, a page-indexed memory architecture for longitudinal context management, and a curated library of composable medical skills. Implemented on top of OpenClaw, an open-source agent orchestration framework, this design provides the basis for an \textit{Agentic Operating System for Hospitals}: a computing layer for coordinating clinical workflows while preserving safety, transparency, and auditability. To evaluate the memory component, we introduce manifest-guided retrieval for hierarchical navigation of longitudinal patient records. In a benchmark derived from the MIMIC-IV dataset (v2.2) comprising 100 de-identified patient records and 300 clinical queries stratified across three difficulty tiers (100 per tier), manifest-guided retrieval matched a metadata-filtered RAG baseline on overall recall (0.877 versus 0.876) while achieving 2.2$\times$ higher precision (0.779 versus 0.352) and retrieving fewer documents; on tier-3 longitudinal queries, manifest recall was 21\% higher (0.846 versus 0.701), confirming that LLM-guided hierarchical navigation is most valuable when queries span multiple care episodes. These results outline a practical path toward hospital-scale agentic infrastructure.
\end{abstract}


\maketitle

\section{Introduction}
Recent work has extended large language models (LLMs) into autonomous agents capable of interacting with external software environments. Agent architectures combine reasoning, planning, tool use, and memory to execute workflows that extend beyond text generation alone \cite{yehudai2025evaluation}. These capabilities have stimulated growing interest in healthcare, where coordination across heterogeneous data sources, clinical documents, and care processes is central to everyday practice.

Several recent studies suggest that LLM-based agents could assist with diagnostic support, clinical documentation, and patient communication \cite{wang2025medicalagents}. Translating such systems into operational hospital settings, however, exposes persistent structural limitations. Existing agentic frameworks generally assume open computing environments and broad operating-system permissions, including unrestricted file-system access, external network calls, and arbitrary code execution. These assumptions are fundamentally misaligned with healthcare privacy regulation and audit requirements. At the same time, dominant memory approaches based on flat vector retrieval fragment patient histories into isolated embeddings, discarding the longitudinal structure on which clinical reasoning depends and weakening recall across extended care episodes.

Hospital environments impose additional constraints. Care is delivered through structured workflows involving multiple participants, including physicians, nurses, and patients. Information exchange occurs primarily through structured artifacts such as clinical notes, laboratory reports, and treatment plans rather than through continuous dialogue. Most current agent architectures, by contrast, assume a single conversational interface rather than a collaborative document-centered environment. A further limitation arises from the design of existing hospital information systems. Electronic health records, clinical decision support tools, and care-coordination platforms typically encode care as fixed, pre-programmed workflows designed for anticipated pathways. These systems perform well for common, well-defined protocols but are poorly suited to the long tail of clinical variability, because patient presentations and disease trajectories rarely conform to predefined routines. Physicians confronting unusual comorbidity profiles, or patients seeking guidance that spans multiple episodes of care, are therefore often left beyond the reach of installed workflows.

Taken together, these observations suggest that adapting agents for healthcare is not primarily a problem of model capability, but of infrastructure design. We argue that agents should operate within a controlled computing layer purpose-built to satisfy the safety, memory, and coordination requirements of clinical environments. We therefore introduce the concept of an \textit{Agentic Operating System for Hospitals}, a computing layer designed to support safe agent deployment in clinical settings. We describe four core components addressing execution control, longitudinal memory organization, document-mediated coordination, and composable clinical skills.
\section{Related Work}

\subsection{LLM agent architectures}

Recent research has explored LLM agents as modular systems integrating reasoning, tool use, and memory.
Architectures such as ReAct \cite{yao2023react}, Reflexion \cite{shinn2023reflexion}, and generative agents \cite{park2023generativeagents} demonstrate how language models can coordinate multi-step interactions with external environments.
These frameworks typically grant agents broad access to system resources, including file systems, external APIs, and arbitrary code execution environments.
While such permissiveness is acceptable in sandboxed research settings, it introduces unacceptable risks in operational environments where safety and auditability are mandatory requirements.
The challenge, therefore, is not solely one of capability but of \textit{containment}: how to embed agentic reasoning within a controlled execution boundary without sacrificing functional utility.
Existing agent frameworks provide no principled solution to this problem, motivating the need for an infrastructure layer analogous to an operating system.
One concrete example is \textit{OpenClaw}, an open-source agent orchestration framework designed for complex, multi-step tasks~\cite{openclaw2025}.
Unlike tools-and-skills integration that many agent frameworks already support, OpenClaw's distinguishing contributions lie in its persistent memory architecture and its heartbeat mechanism: a scheduled task system that enables agents to autonomously initiate actions at defined intervals without requiring an explicit user trigger.
Together, these two mechanisms allow agents to maintain coherent long-term context across sessions and to proactively monitor or act upon evolving conditions---capabilities that are essential for any system expected to operate continuously in a dynamic environment.
However, OpenClaw's design targets a permissive, consumer-facing context and therefore lacks the execution isolation, audit trail mechanisms, and clinically structured longitudinal memory required for hospital deployment.

\subsection{Retrieval-augmented generation in clinical contexts}

Retrieval-augmented generation (RAG) systems have been proposed to address the limitations of static model knowledge by integrating external document repositories \cite{lewis2020rag}.
By retrieving relevant passages at inference time, RAG enables models to reason over knowledge that would otherwise fall outside their training distribution.
However, conventional RAG pipelines rely on flat vector similarity retrieval over fragmented text chunks, producing ranked lists of decontextualized passages that may span unrelated encounters, time points, or clinical episodes.
A query about a patient's medication history may surface a dosage note from three years prior alongside a recent adverse-event report, with no structural signal to distinguish their temporal or causal relationship.
A patient record is not a collection of independent sentences; it is an ordered sequence of temporally linked documents---admission notes, progress notes, laboratory reports, imaging findings, and discharge summaries---whose meaning depends on sequential context.
Chunking such records into isolated embeddings sacrifices precisely the contextual dependencies that clinical reasoning requires.

Several approaches have attempted to address structural limitations in RAG.
MemGPT \cite{packer2023memgpt} introduces a hierarchical memory model inspired by OS virtual memory paging, moving context between fast in-context storage and slower external storage; however, its paging policy is designed for conversational continuity rather than document-structured clinical knowledge.
GraphRAG \cite{edge2024graphrag} constructs a knowledge graph over a corpus and retrieves via community summaries, improving multi-hop reasoning but requiring expensive offline graph construction that does not accommodate the continuous document mutations characteristic of live patient records.
RAPTOR \cite{sarthi2024raptor} builds a recursive tree of abstractive summaries for hierarchical retrieval, but operates over static corpora and provides no mechanism for incremental index maintenance on document updates.
HippoRAG \cite{gutierrez2024hipporag} models retrieval as associative memory over a knowledge graph, improving cross-document reasoning, but likewise assumes a static document collection.
By contrast, we propose a page-indexed memory architecture that abandons vector similarity entirely.
Rather than encoding documents as embeddings, it exposes the document hierarchy to the agent through human-readable manifest files and relies on the agent's own language-reasoning capability to navigate progressively toward relevant content.
This design eliminates embedding computation overhead, requires no offline index construction, and naturally handles live document mutations without any reindexing step.

\subsection{Clinical workflow integration and deployment challenges}

Within healthcare, substantial work has examined the requirements for integrating AI systems into clinical environments \cite{sendak2020clinicalworkflow, nagendran2020aiworkflow, rajpurkar2022healthcareai}.
These studies consistently show that successful deployment depends on alignment with existing institutional structures---documentation practices, professional role boundaries, and regulatory governance---rather than with the open, permissive computing assumptions that underlie general-purpose agent frameworks.
A distinctive feature of hospital environments is their inherently multi-participant structure: physicians, nurses, and patients each occupy defined roles and exchange information primarily through structured documents rather than continuous dialogue.
This stands in sharp contrast to general-purpose agent architectures, which assume a single conversational interface between one user and one model instance.
Clinical settings therefore demand a multi-agent architecture in which separate agents serve distinct participant roles and coordinate indirectly through shared document artifacts.
Regulatory requirements in healthcare further demand that system behaviour be explainable and auditable, placing additional constraints on memory design, skill execution, and inter-agent communication.

A deeper and more persistent limitation, however, concerns the design philosophy of hospital information systems as a class.
The full ecosystem of clinical software---including hospital information systems (HIS), electronic health records (EHR), clinical decision support systems (CDSS), triage platforms, pharmacy and drug management systems, and care coordination tools---is built on a common principle: each system implements a fixed, pre-programmed set of functions, and any clinical task not explicitly anticipated at design time falls outside the system's scope.
This design philosophy produces systems that are reliable and auditable within their predefined boundaries but structurally incapable of responding to ad-hoc clinical needs \cite{ash2004consequences, sutton2020cdss}.
Because no two patients present identically and no disease trajectory follows a scripted sequence, the diversity of real-world clinical requests systematically exceeds what any fixed function library can represent.
A physician managing a patient with an unusual comorbidity combination, a triage nurse assessing a presentation that spans multiple protocol categories, or a patient seeking guidance that crosses the boundary between drug management and chronic disease monitoring will find that no installed system can serve the request in its entirety.
This gap is not incidental but structural: fixed-function systems cannot reason about novel situations, only match incoming requests against predefined rules \cite{panch2019inconvenient}.
The introduction of agentic AI---systems that can compose ad-hoc task sequences from a library of fine-grained capabilities and build longitudinal context through long-term memory---offers a qualitatively different approach to this problem, one that can address the long tail of clinical need rather than serving only its modal case.

\subsection{Research gaps}

Taken together, the literature reveals four unresolved gaps.
First, current agent frameworks lack a principled mechanism for restricting execution to well-defined, auditable actions.
Second, existing memory systems based on flat vector retrieval are poorly suited to the hierarchical, longitudinal structure of clinical records.
Third, no existing architecture explicitly models the multi-participant, document-mediated nature of hospital workflows.
Fourth, existing hospital IT systems implement care as static, pre-programmed workflows that cannot adapt to the inherent long-tail variability of clinical practice; physicians and patients regularly encounter situations that fall outside installed protocols, with no mechanism for ad-hoc task execution.
Our work addresses each of these gaps directly: a restricted execution environment inspired by Linux multi-user systems targets the first; a page-indexed memory architecture addresses the second; a document-centric multi-agent design addresses the third; and the agentic task-composition model---through which the agent reasons over a library of medical skills and composes them dynamically to address novel clinical requests---targets the fourth.
Together these components constitute the proposed \textit{Agentic Operating System for Hospitals}.

\section{System Architecture}

\subsection{Restricted execution and skills coordination}

The proposed architecture treats the hospital environment as a first-class design constraint and structures agent deployment accordingly.
It builds upon the skill-based decomposition principle of OpenClaw while replacing its permissive runtime with Linux-based execution isolation and adding the longitudinal memory mechanisms that clinical deployment requires.
The central design principle is \textit{least-privilege execution}: rather than granting agents ambient access to system resources, each agent instance operates within an isolated runtime whose permission boundary is statically defined at deployment time, analogous to a restricted user account in a Unix multi-user system.

Each participant role---patient, physician, or clinical staff---is mapped to a dedicated agent process executing within its own isolated namespace.
The namespace exposes a fixed interface consisting exclusively of sanctioned tool invocations; direct file-system access, outbound network connections, and dynamic code loading are prohibited at the runtime level rather than enforced through model-level instruction.
This design shifts trust enforcement from the model's instruction-following behaviour to the execution environment itself, providing stronger and more auditable safety guarantees.

Agent capabilities are realized through a curated library of \textit{medical skills}: statically defined, pre-audited workflow modules that encapsulate clinical operations such as vital-sign aggregation, medication adherence tracking, and structured report generation.
Each skill presents a typed interface to the agent and is permitted to access only internal, institutionally managed resources---such as a hospital's laboratory information system, pharmacy database, or wearable sensor gateway---through predefined, narrowly scoped connectors; no skill may invoke arbitrary external network calls, escalate its own privileges, or execute arbitrary system calls.
This design preserves the network-isolation property of the agent runtime while still enabling skills to interact with real clinical data sources through auditable, pre-approved integration points.
This skill-level decomposition enables formal pre-deployment review and fine-grained runtime monitoring of all agent-initiated actions.

A defining capability that distinguishes this architecture from conventional hospital IT systems is its support for \textit{ad-hoc task composition}.
Existing hospital information systems encode care as static, pre-programmed workflows: each system function is defined at deployment time, and requests that fall outside those definitions are unserviceable.
The agentic model inverts this constraint.
Because the agent can reason over the available skill library and dynamically compose sequences of skill invocations in response to a stated clinical goal, it can address novel situations that no pre-existing workflow covers.
A physician who needs to correlate a rare drug interaction with multi-year laboratory trends, or a patient who wants to understand how a new diagnosis relates to their existing comorbidity history, poses a request that no fixed workflow can serve; the agent constructs a task plan on demand, navigates the manifest hierarchy to retrieve the relevant longitudinal context, and invokes the appropriate sequence of skills to produce a coherent clinical response.
The long-term memory architecture reinforces this capability: because the agent accumulates context across care episodes through the document hierarchy rather than resetting between interactions, it can identify patient-specific patterns and address personalized clinical needs that emerge only over extended longitudinal observation---the class of need that is precisely most underserved by pre-programmed hospital systems.

Inter-agent coordination follows a document-mutation model rather than a shared-memory or direct-message model.
All information exchange is realized as structured writes to shared clinical documents; no agent communicates with another through a direct channel.
Patient agents hold write access to patient-owned documents such as symptom logs, vital-sign records, and self-reported health journals, while clinician agents hold write access to physician-authored documents such as assessment notes, medication orders, and follow-up plans.
Cross-role interactions are mediated exclusively by these documents: a patient agent that detects an out-of-range vital-sign appends a structured alert entry to the patient record, which the clinician agent then reads; a clinician agent that revises a follow-up schedule writes the updated plan as a new document version, which the patient agent subsequently ingests to adjust its monitoring behaviour.

\begin{figure}[htbp]
\centering
\includegraphics[width=0.95\linewidth]{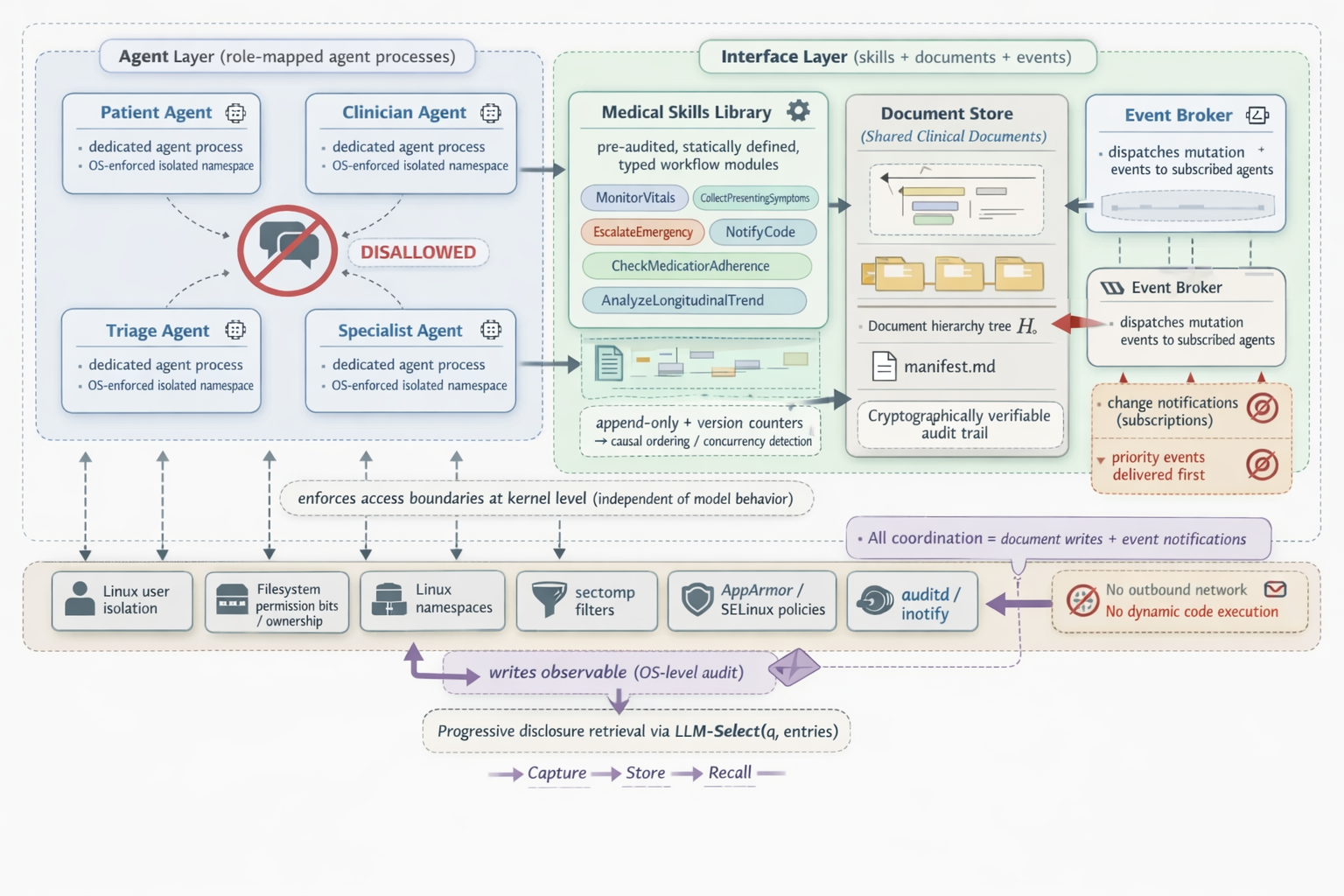}
\caption{Architecture of the Agentic Operating System for Hospitals.
The system comprises three layers.
(\textit{Agent Layer}) Each participant role---patient, clinician, triage, specialist---maps to a dedicated agent process executing within an OS-enforced isolated namespace; agents interact with the environment solely through two operations: invoking pre-audited medical skills and reading or writing shared clinical documents.
(\textit{Interface Layer}) The Medical Skills Library exposes a curated set of statically typed workflow modules; the Document Store maintains per-role document collections linked by an append-only mutation event stream dispatched by the Event Broker.
(\textit{OS Enforcement Layer}) Linux user isolation, file permission bits, \texttt{seccomp} filters, AppArmor policies, and the \texttt{auditd}/\texttt{inotify} subsystem collectively enforce all access boundaries at the kernel level, independent of model behaviour.
Critically, no agent communicates with another through a direct channel; all inter-agent coordination is realised exclusively through document writes and event notifications.}
\label{fig:arch}
\end{figure}

To support reactive coordination, each agent subscribes to a change-notification stream associated with the documents within its access scope.
Internally, every document write is recorded as an append-only mutation event containing a document identifier, the modified page reference, the writing agent's role, and a monotonically increasing version counter.
A lightweight event broker dispatches these mutation events to all subscribed agents; upon receipt, an agent evaluates whether the change warrants immediate action---such as escalating an abnormal finding to a clinician---or deferred processing during the next scheduled reasoning cycle.
This subscription mechanism decouples agents temporally while preserving causal ordering: because every event references a version counter, agents can reconstruct the exact sequence of document changes and detect concurrent modifications without polling.
To prevent logically inconsistent concurrent updates, write-capable reasoning cycles are serialized at the patient scope through a short-lived filesystem lease.
Before an agent begins a read--reason--write operation over a patient's document subtree, it must first acquire that patient's lock directory using an atomic filesystem primitive such as \texttt{mkdir} or \texttt{open(O\_CREAT|O\_EXCL)}.
The lease record stores the agent identity, acquisition time, and last-heartbeat timestamp; the agent renews the lease while running and releases it immediately after the final page append completes.
If the agent crashes, stalls, or exceeds a bounded execution window, the lease expires automatically and can be safely reclaimed after a timeout.
This policy intentionally chooses coarse-grained locking: all writes affecting a single patient are serialized for a short interval, while unrelated patients remain fully parallelizable across the system.
Because the mechanism is implemented directly through standard Linux filesystem semantics, it requires no distributed lock service and remains fully auditable through the same OS-native monitoring stack as ordinary document writes.
All mutation events are persisted alongside the document itself, yielding a complete, tamper-evident audit trail of every inter-agent information exchange across the system; Figure~\ref{fig:arch} summarises the overall architecture.
Among the four components described above, the page-indexed memory architecture requires the most detailed treatment and is elaborated in the following subsection.

\subsection{Page-indexed memory architecture}

\begin{figure}[htbp]
\centering
\includegraphics[width=0.95\linewidth]{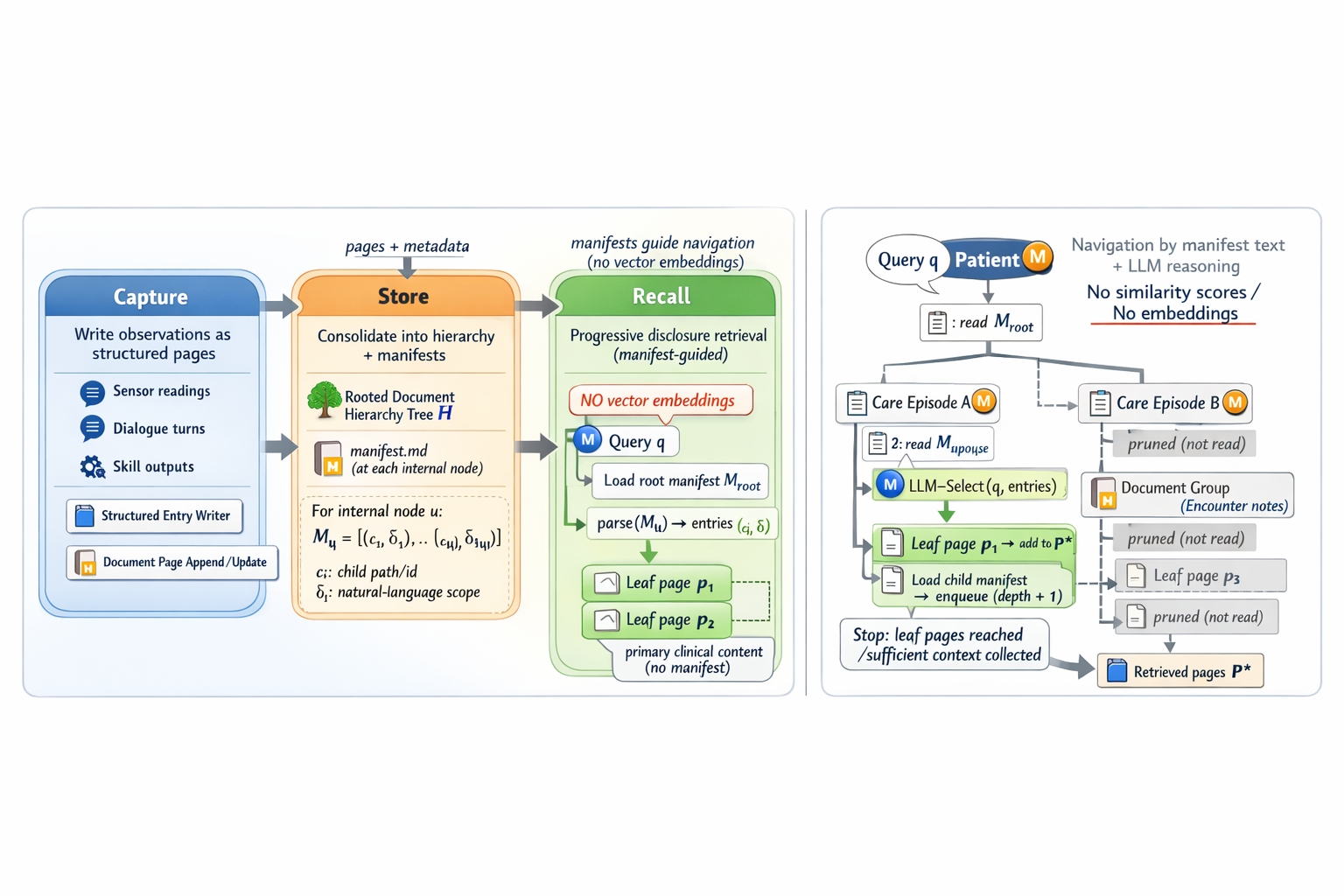}
\caption{Page-indexed memory architecture and progressive disclosure retrieval.
(\textit{Left}) The memory lifecycle consists of three stages: \textit{Capture} (observations are written as structured pages), \textit{Store} (pages are consolidated into a rooted document hierarchy with a manifest file at each internal node), and \textit{Recall} (the agent navigates the hierarchy through successive manifest reads).
(\textit{Right}) The document tree is organised by patient, care episode, and document group; each internal node carries a manifest (badge~\textbf{M}) listing its children with natural-language scope descriptions, while leaf nodes contain primary clinical content.
Given a query, the agent issues an LLM-Select call at each level to identify relevant subtrees (highlighted in green) and prune irrelevant branches without reading them (shown in grey, labelled ``pruned'').
This design requires no vector embeddings; retrieval accuracy depends solely on the agent's language-reasoning capability applied to manifest text, and the sequence of manifests consulted constitutes an inherently interpretable access trace.}
\label{fig:memory}
\end{figure}

The page-indexed memory architecture organizes agent memory as a three-stage lifecycle: \textit{capture}, \textit{store}, and \textit{recall} (Figure~\ref{fig:memory}).
In the capture stage, observations produced during agent interactions---sensor readings, dialogue turns, skill outputs---are written as structured entries into the document collection via the mutation mechanism described above.
In the store stage, the system consolidates captured entries into a document hierarchy and maintains a \textit{manifest} at each hierarchical level to summarize its contents.
In the recall stage, the agent navigates this hierarchy through successive manifest reads, progressively narrowing its focus until it reaches the pages most relevant to its current reasoning task.
Critically, this architecture requires no vector embeddings: navigation is driven entirely by the agent's own language-reasoning capability applied to manifest text.

\paragraph{Document hierarchy and manifest files.}

The document collection is organized as a rooted tree $\mathcal{H}$.
Internal nodes of $\mathcal{H}$ represent document groups (e.g., all records for a given patient, or all encounters within a care episode), and leaf nodes are individual content pages $p$ containing clinical content such as encounter notes, laboratory results, or medication orders.
Each internal node $u$ stores a \texttt{manifest.md} file listing its immediate children with a brief, human-readable description of each child's scope and contents.
Formally, the manifest at node $u$ is a structured document:
\[
M_u = \bigl[(c_1,\, \delta_1),\; (c_2,\, \delta_2),\; \ldots,\; (c_{|u|},\, \delta_{|u|})\bigr]
\]
where $c_i$ is the identifier (path) of the $i$-th child and $\delta_i$ is a concise natural-language description of that child's clinical scope, time range, and document type.
Leaf pages carry no manifest; they contain only primary content.

Manifests are generated and maintained by the system automatically: when a new page is appended to a document group following a mutation event, the system issues an LLM call to produce or update $\delta_i$ for the affected child entry.
Because each manifest update is local to a single node, maintenance cost is $O(d)$ per mutation event, where $d$ is the depth of the modified node in $\mathcal{H}$, and does not require batch reprocessing of sibling nodes.

\paragraph{Progressive disclosure retrieval.}

Rather than computing similarity scores, the agent retrieves relevant content by iteratively reading manifests and selecting which subtrees to descend into, guided by its reasoning about the query.
This \textit{progressive disclosure} strategy mirrors how a clinician navigates a paper chart: first scanning a summary index, then opening only the sections likely to contain the answer.

Algorithm~\ref{alg:retrieval} formalizes this process.
The agent begins at the root manifest $M_\text{root}$ and at each level presents the manifest entries to the LLM together with the current query; the LLM selects a subset of children to expand.
Branches not selected are pruned without being read, avoiding unnecessary context consumption.
Traversal continues until either leaf pages are reached or the agent determines it has collected sufficient context.

\begin{algorithm}
\caption{Progressive Disclosure Retrieval via Manifest Navigation}
\label{alg:retrieval}
\begin{algorithmic}[1]
\STATE \textbf{Input:} query $q$, root manifest $M_\text{root}$, max depth $L$
\STATE \textbf{Output:} retrieved pages $P^*$
\STATE $P^* \leftarrow \emptyset$
\STATE $\text{frontier} \leftarrow \{(M_\text{root},\; \text{depth}=0)\}$
\WHILE{$\text{frontier} \neq \emptyset$}
  \STATE $(M_u,\; \ell) \leftarrow \text{dequeue}(\text{frontier})$
  \STATE $\text{entries} \leftarrow \text{parse}(M_u)$ \COMMENT{list of $(c_i, \delta_i)$ pairs}
  \STATE $S \leftarrow \text{LLM-Select}(q,\; \text{entries})$ \COMMENT{agent selects relevant children}
  \FOR{each $(c_i, \delta_i) \in S$}
    \IF{$c_i$ is a leaf page}
      \STATE $P^* \leftarrow P^* \cup \{\text{load}(c_i)\}$
    \ELSIF{$\ell < L$}
      \STATE enqueue $(\text{load manifest of } c_i,\; \ell + 1)$ into frontier
    \ELSE
      \STATE $P^* \leftarrow P^* \cup \{\text{load}(c_i)\}$ \COMMENT{depth limit reached; load node content as partial context}
    \ENDIF
  \ENDFOR
\ENDWHILE
\STATE \textbf{return} $P^*$
\end{algorithmic}
\end{algorithm}

The \textsc{LLM-Select} call at line~8 is the sole decision-making step: the agent reads the manifest entries as natural language and outputs the subset of child identifiers relevant to the query.
This replaces the similarity function of embedding-based approaches with the agent's full language-understanding capability, allowing it to reason about clinical scope, temporal relevance, and document type simultaneously rather than reducing all three dimensions to a scalar distance.

The depth bound $L$ limits the maximum number of manifest reads per query and controls the breadth-depth trade-off: shallow $L$ retrieves broader context from higher-level summaries, while deeper traversal reaches more specific page content.
In practice, $L$ is set per query type: a broad clinical summary query terminates at depth~1 or~2, while a targeted medication lookup descends to leaf pages.

\paragraph{Manifest maintenance on document mutation.}

Each document mutation event (described above) may invalidate the manifest of the parent node.
Upon receiving a mutation event for page $p$ under node $u$, the system issues an incremental LLM call that rewrites only the affected child entry $\delta_i$ in $M_u$, leaving all other entries unchanged.
If a page is inserted or deleted, the corresponding entry is added or removed from $M_u$ and the description regenerated for adjacent entries if their scope changes.
Manifests at ancestor nodes are updated only when the scope description of the modified subtree changes materially (e.g., a new clinical episode begins), determined by a brief LLM comparison of the old and new $\delta_i$.
This localized update policy ensures that manifest files remain accurate reflections of live document state at the cost of at most $O(L)$ incremental LLM calls per mutation event, with no embedding recomputation at any level.

\subsection{Clinical workflow integration}

\begin{figure}[htbp]
\centering
\includegraphics[width=0.95\linewidth]{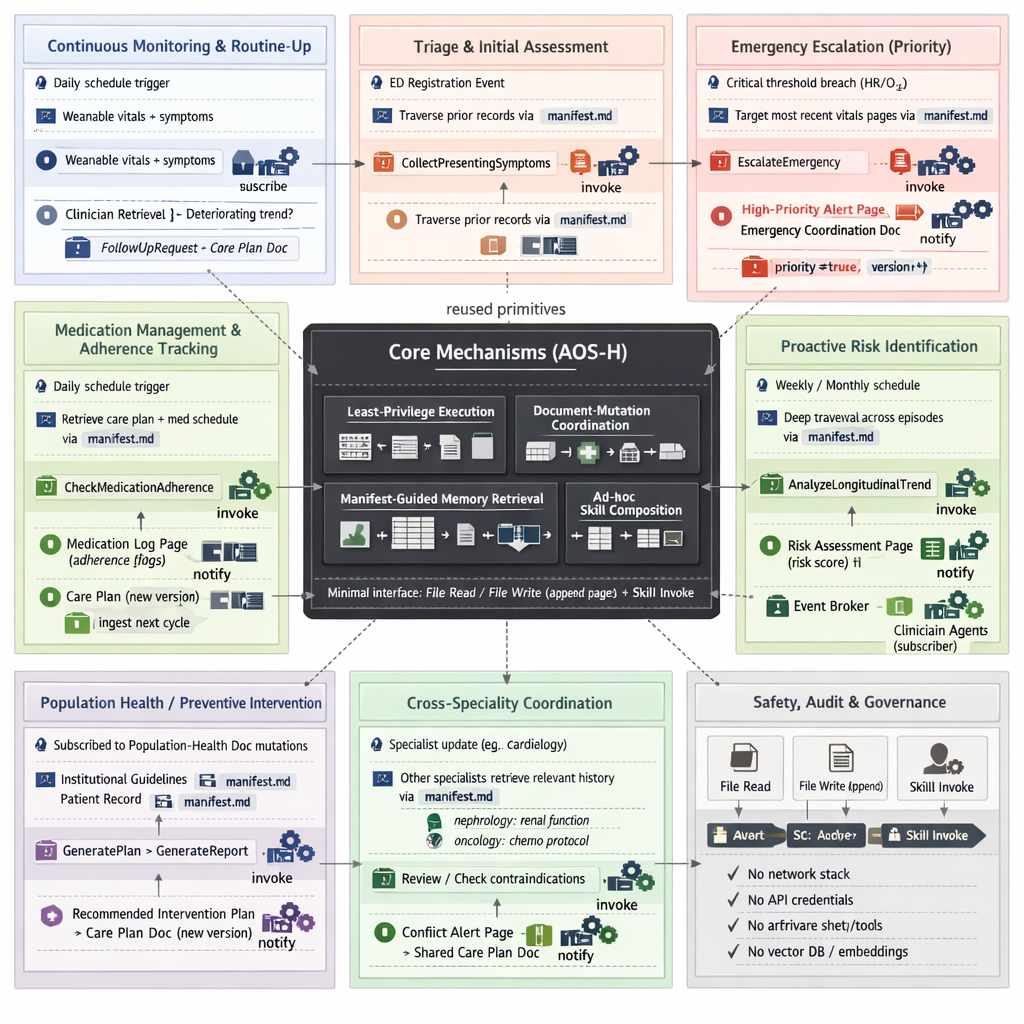}
\caption{Clinical application scenarios enabled by the Agentic Operating System for Hospitals.
The nine-panel grid illustrates representative use cases realised by the proposed architecture.
The central panel summarises the four core mechanisms shared across all scenarios: least-privilege execution, document-mutation coordination, manifest-guided memory retrieval, and ad-hoc skill composition.
The surrounding eight panels cover scheduled care (\textit{Continuous Monitoring}, \textit{Medication Management}), reactive workflows (\textit{Triage}, \textit{Emergency Escalation}), longitudinal analysis (\textit{Proactive Risk Identification}, \textit{Population Health}), multi-agent coordination (\textit{Cross-Specialty Coordination}), and infrastructure (\textit{Safety, Audit \& Governance}).
Each scenario is realised entirely through the composition of medical skills, document writes, event subscriptions, and manifest navigation---without any bespoke architectural extension.
The diversity of use cases demonstrates that a single unified infrastructure layer is sufficient to address both routine and exceptional clinical needs across the full spectrum of hospital workflows.}
\label{fig:apps}
\end{figure}

Hospitals involve numerous structured processes that require coordinated action across time, participants, and information sources.
The proposed architecture supports these processes by composing three mechanisms introduced above: medical skills as the unit of executable action, document mutation as the communication channel between agents, and manifest-guided memory navigation as the means of retrieving relevant clinical context.
The following scenarios illustrate how representative hospital workflows are realized within this framework (Figure~\ref{fig:apps}).

\paragraph{Continuous monitoring and routine follow-up.}

A patient agent assigned to a post-discharge patient executes a \textsc{MonitorVitals} skill on a scheduled interval, aggregating wearable sensor readings and patient-reported symptoms into a structured daily summary page appended to the patient's health record.
When the clinician agent's next scheduled reasoning cycle fires, it invokes manifest-guided retrieval to collect recent summary pages, synthesizes a progress assessment, and appends a follow-up note to the physician document.
If the assessment identifies a deteriorating trend---such as progressively elevated blood pressure over five consecutive days---the clinician agent appends a structured \textsc{FollowUpRequest} entry to the shared care plan document, triggering a mutation event that notifies a scheduling agent---a dedicated agent process responsible for managing appointment queues and booking workflows---to book an in-person review.
Throughout this workflow no agent communicates directly with another; all coordination is mediated through document writes observed via the event subscription model.

\paragraph{Triage and initial assessment.}

When a patient registers at the emergency department, a triage agent executes a \textsc{CollectPresentingSymptoms} skill to populate the initial encounter page of the patient record.
The agent then performs manifest-guided traversal of prior records---navigating episode-level manifests to retrieve relevant chronic conditions, current medications, and recent laboratory results---and appends a structured triage assessment page consolidating the presenting complaint, relevant history, and a preliminary acuity score.
This assessment page is written to a shared triage queue document accessible to all on-duty clinician agents.
A clinician agent subscribed to that queue receives the mutation event, reads the triage page, and either accepts the case or re-routes it to a specialist agent by writing a referral entry to the queue document, updating the patient's acuity manifest entry accordingly.

\paragraph{Emergency escalation.}

When a patient agent detects a critical physiological event---such as a heart-rate reading exceeding a predefined threshold or an oxygen-saturation drop below a safe boundary---it immediately invokes the \textsc{EscalateEmergency} skill rather than deferring to the next scheduled reasoning cycle.
This skill atomically appends a high-priority alert page to the patient record and writes a structured escalation entry to a dedicated emergency coordination document monitored by all on-duty clinician agents.
Because escalation entries carry a priority flag in the mutation event payload, the event broker delivers them ahead of standard notifications, allowing clinician agents to interrupt their current reasoning cycle and respond within a bounded latency.
The responding clinician agent reads the patient record via manifest navigation---directly targeting the most recent vital-sign pages---documents an initial response plan, and invokes a \textsc{NotifyCode} skill that triggers the appropriate hospital alert protocol.
Every step in this chain---sensor event, escalation write, clinician response, and protocol invocation---is persisted as a timestamped mutation event, producing a complete and auditable timeline of the emergency response.

\paragraph{Medication management and adherence tracking.}

A patient agent executes a \textsc{CheckMedicationAdherence} skill daily, comparing scheduled medication events in the care plan against patient-confirmed administration entries.
If a missed dose is detected, the agent appends an adherence flag to the medication log page; if the pattern persists beyond a configurable threshold, it writes a non-adherence alert to the shared care coordination document.
The clinician agent, upon receiving the mutation notification, retrieves the medication history pages via manifest navigation and determines whether to adjust the dosing schedule, initiate a patient-education interaction, or escalate to a pharmacist agent.
Any modification to the care plan is written as a new versioned page, and the patient agent ingests the change at its next cycle to update its monitoring parameters accordingly.
This closed-loop design ensures that medication management remains consistent across agent boundaries without requiring any direct inter-agent message passing.

\paragraph{Proactive risk identification and preventive intervention.}

Beyond reacting to acute threshold breaches, the architecture naturally supports proactive longitudinal reasoning by exploiting the full temporal depth of the document hierarchy.
On a weekly or monthly schedule, a clinician agent executes an \textsc{AnalyzeLongitudinalTrend} skill that issues a broad manifest-guided retrieval query spanning multiple care episodes---navigating from the patient's root manifest across years of encounter summaries, laboratory trend pages, and vital-sign history.
The agent reasons over this longitudinal context to compute a structured risk assessment page, recording indicators such as a gradual decline in renal function across twelve months of laboratory results or a pattern of increasing emergency visits preceding each winter.
This risk assessment page is appended to the patient record and, if the computed risk score exceeds a configurable threshold, additionally written as a preventive alert to a population-health coordination document monitored by a care management agent.
The care management agent, upon receiving the mutation event, reviews the alert, cross-references institutional preventive care guidelines via its own manifest-guided retrieval, and writes a recommended intervention plan---such as a nephrology referral or a structured self-management programme---to the patient's care plan document.
The patient agent ingests the updated care plan and adjusts its monitoring focus accordingly, increasing the frequency of relevant skill executions.
This workflow requires no architectural extension; it is realized entirely through scheduled skill invocations, manifest-guided retrieval across deep document history, and the existing document-mutation coordination model.

\paragraph{Cross-specialty coordination for multimorbid patients.}

Patients managed concurrently by multiple specialists present a coordination challenge that the document-mutation model addresses directly through shared document access and event subscription.
Each specialist discipline is represented by a dedicated clinician agent---for example, a cardiology agent, a nephrology agent, or an oncology agent---each holding write access to its own specialty-note documents and read access to a shared multidisciplinary care plan document.
When the cardiology agent updates the antihypertensive regimen, it writes a new medication order page and appends a summary entry to the shared care plan; the resulting mutation events are delivered to all subscribed specialist agents.
Each specialist agent that receives the notification performs a manifest-guided retrieval of its own relevant history---the nephrology agent navigates to recent renal function pages, the oncology agent to current chemotherapy protocol pages---and evaluates the change against its own treatment context.
If a conflict is identified, such as a newly prescribed ACE inhibitor contraindicated by the patient's current chemotherapy agent, the detecting agent appends a structured conflict alert page to the shared care plan document, triggering a further mutation event visible to all specialist agents and the coordinating clinician.
The coordinating clinician agent aggregates these conflict entries, retrieves the full relevant context across all specialty manifests, and produces a reconciled care plan page that supersedes the conflicting orders.
Because every write in this workflow is a versioned document mutation, the full sequence of specialist updates, conflict detections, and plan reconciliations is preserved in the audit trail with causal ordering intact, providing a transparent record equivalent to a documented multidisciplinary team meeting.

\subsection{Safety and governance}

A key insight of the proposed architecture is that reducing the agent's effective interface to two primitive operations---file read and file write---allows most of the safety and governance problem to be delegated to the security machinery of the host operating system rather than implemented as bespoke application logic.

\paragraph{Minimal trusted computing base.}

Unlike general-purpose agent frameworks that require network access, dynamic code execution, database connections, and external API calls---each of which introduces an independent attack surface---agents in the proposed system interact with the environment solely by reading documents and appending pages.
This minimal interface dramatically shrinks the trusted computing base: the system has no network stack to secure, no interpreter sandbox to maintain, and no API credential lifecycle to manage.
The security problem reduces to controlling which agent processes may read or write which files, a problem that mature operating systems have solved for decades.

\paragraph{OS-native enforcement.}

On Linux, the proposed permission model maps directly onto standard OS security primitives without any additional infrastructure.
Each agent process runs under a dedicated OS user account; file ownership and permission bits enforce the read/write access boundaries defined by the architecture---patient agents own patient-record files, clinician agents own physician-document files, and no agent can access files outside its assigned subtree.
Mandatory access control frameworks such as SELinux or AppArmor can further constrain each agent process to its declared file paths at the kernel level, preventing privilege escalation even if the model's instruction-following is compromised.
Linux namespaces provide filesystem-level isolation between agent processes at no additional development cost, and \texttt{seccomp} filters can restrict the system-call surface of each process to the minimal set required for file I/O.
Because these mechanisms are enforced by the kernel rather than by application code, they are independent of model behaviour and cannot be circumvented by a misbehaving or adversarially prompted agent.

\paragraph{Audit trail as a filesystem property.}

The append-only mutation log that underlies inter-agent coordination (Section~3) is, at the storage layer, an ordinary append-only file.
Filesystem-level auditing tools such as Linux \texttt{inotify} and \texttt{auditd} can therefore monitor every document write in real time without any instrumentation of agent code.
This means that a complete audit trail of all agent actions---required by healthcare regulations such as HIPAA---is obtained as a byproduct of the OS's native file-audit infrastructure rather than as a separately engineered component; tamper-evidence can be further strengthened by routing the append-only log through a hash-chaining mechanism at the storage layer.

\paragraph{Comparison with complex agent architectures.}

The safety simplification achieved here contrasts sharply with approaches that grant agents broad system access and then attempt to impose safety constraints on top.
Securing an agent that can invoke arbitrary APIs, execute shell commands, and load external libraries requires custom sandboxing, runtime policy enforcement, and continuous monitoring of a large and evolving attack surface.
By contrast, securing an agent whose only interface is file I/O requires configuring OS users, file permissions, and an access-control policy---tasks that system administrators perform routinely and for which well-understood tooling already exists.
This alignment with existing OS security infrastructure not only reduces implementation complexity but also facilitates independent security audits, as reviewers can verify the permission model using standard OS introspection tools rather than inspecting novel application-layer enforcement code \cite{holzinger2022trustworthy}.

\section{Evaluation}

We conducted a targeted empirical evaluation of the page-indexed memory component. 
The goal was to test whether manifest-guided progressive disclosure recovers relevant evidence more reliably than flat vector retrieval on structured longitudinal clinical records.

\subsection{Experimental setup}

\paragraph{Data source and patient selection.}
We evaluated on MIMIC-IV v2.2~\cite{johnson2023mimiciv}, a de-identified electronic health record dataset from Beth Israel Deaconess Medical Center, supplemented with clinical notes from MIMIC-IV-Note~\cite{johnson2023mimicivnote} (available through PhysioNet~\cite{goldberger2000physionet}).
Candidate patients were required to have at least three hospital admissions and to pass a quality floor: discharge-note coverage ${\geq}50\%$ of admissions, diagnosis diversity ${\geq}5$ unique ICD codes, at least one radiology report, and de-identification density ${\leq}15$ tokens per 1\,000 characters.
From eligible candidates we drew a stratified random sample of 100 patients across gender, age bracket ($<$50, 50--70, $>$70), and admission-count bracket (3--5, 6--9, $\geq$10).

\paragraph{Corpus construction.}
MIMIC-IV stores clinical data as relational tables in compressed CSV files.
A deterministic pipeline reorganized these flat tables into a three-level directory hierarchy---patient (\texttt{S\{id\}})~/~admission (\texttt{hadm\_\{id\}})~/~document---mirroring the structure required by the manifest architecture.
Six document types were extracted per admission where available: discharge summaries, radiology reports, ICD-coded diagnoses, emergency-department triage assessments, prescriptions, and laboratory results.
Free-text notes (discharge summaries and radiology reports) were preserved verbatim, retaining only the highest-sequence revision for discharge notes.
Radiology reports were sorted by chart time and numbered sequentially within each admission.
Structured data were rendered into human-readable Markdown: diagnoses were joined with ICD long titles, prescriptions were sorted chronologically, laboratory results were grouped by test category (e.g.\ Chemistry, Hematology) using the \texttt{d\_labitems} dictionary, and triage records were formatted as vital-sign summaries.
All documents received YAML provenance headers recording source table, identifiers, and metadata.
No clinical content was added or paraphrased; the pipeline only restructured, formatted, and annotated original records.
The resulting corpus contains 6\,991 leaf documents across 814 admissions.

To enable hierarchical navigation, manifests were generated bottom-up by the LLM as Markdown tables with four columns (Child, Type, Date, Scope).
Admission-level manifests were produced from the admission's metadata and the first 1\,500 characters of each leaf document, with the Scope column summarizing each document's clinical content in one sentence.
Patient-level manifests were then produced from each patient's demographics and a structured summary per admission (dates, document count, top three ICD diagnoses), with Scope summarizing the admission's key events.
The 914 manifests serve as the navigational index for the retrieval system.

\paragraph{Queries and gold labels.}
Queries were generated from the selected cohort using an LLM-guided pipeline over the manifest hierarchy. For each patient, feasible tiers were inferred from corpus structure: Tier~1 required at least one leaf document, Tier~2 required at least one admission with two or more distinct document-type categories, and Tier~3 required at least two admissions containing documents. Each patient was then assigned one feasible tier by greedy deficit balancing. The LLM received the patient-level manifest, all admission-level manifests, and snippets from up to five leaf documents per admission (the first 3\,000 characters of each, prioritized by document type and explicitly including laboratory results when available), and produced one candidate query together with provisional gold-document paths. The prompt enforced tier-specific constraints: Tier~1 queries had to be answerable from exactly one leaf document, Tier~2 queries from two to three documents within a single admission, and Tier~3 queries from documents spanning at least two admissions. Proposed paths were normalized against the on-disk corpus and discarded if unresolvable; when a tier remained below quota, additional queries were generated from patients feasible for that tier until 100 queries per tier were reached. This process yielded 300 evaluation queries in total, balanced evenly across the three difficulty tiers.

Gold-standard evidence sets and reference answers were constructed through a three-stage LLM-assisted pipeline. First, a \emph{sufficiency check} determined whether the provisional gold documents were enough to answer the query. If not, an \emph{expansion step} used the target patient's admission-level manifests to identify additional documents from their scope descriptions, after which the expanded set was re-verified. Queries whose evidence sets remained insufficient were manually reviewed against the source corpus and corrected before benchmarking. Finally, a concise reference answer was generated using only the verified gold documents.

\paragraph{Models and parameters.}
All LLM calls---manifest navigation, answer generation, gold-label production, and judging---used Kimi K2.5 through an OpenAI-compatible API from Alibaba Cloud.
Manifest navigation and the LLM judge used \texttt{max\_tokens}${}=768$; answer generation for all three systems used \texttt{max\_tokens}${}=2048$.

\paragraph{Systems compared.}
All three systems operated in a patient-scoped setting: patient identity was resolved deterministically from the query text or the structured \texttt{target\_patient} field, without an LLM call.
\textit{Manifest-guided retrieval} then navigated two levels of the hierarchy via LLM calls---first selecting admissions from the patient-level manifest (level~1, hereafter L1), then selecting documents from each admission-level manifest (level~2, L2)---to assemble a targeted set of leaf documents.
\textit{Flat-RAG retrieval} applied section-aware chunking (splitting on Markdown headings, falling back to paragraph boundaries, with a 512-token sliding window and 64-token overlap for oversized sections), embedded each chunk using \texttt{text-embedding-v4} (1\,536 dimensions, Alibaba Cloud), and performed document-level aggregation: $3k$ candidate chunks were grouped by source document, ranked by a composite key (best chunk distance, supporting chunk count, mean top-3 distance), and the top-$k$ \textit{documents} were selected.
\textit{Metadata-filtered RAG} (RAG-AF) applied the same embedding and aggregation pipeline but pre-filtered chunks by admission identifiers extracted from the query text before dense retrieval, falling back to patient-level filtering when no admission identifiers were found. This variant tests whether explicit structural metadata narrows the gap with manifest-guided retrieval.
We report $k{=}10$ as the primary baseline; $k{=}5$ and $k{=}15$ appear in Table~\ref{tab:rag_k}.

\paragraph{Metrics.}
Retrieval precision $= |\mathcal{R} \cap \mathcal{G}| / |\mathcal{R}|$ and recall $= |\mathcal{R} \cap \mathcal{G}| / |\mathcal{G}|$, where $\mathcal{R}$ is the set of retrieved documents and $\mathcal{G}$ the verified gold evidence set.
We also report \textit{perfect recall} (fraction of queries recovering all gold documents) and the mean number of retrieved documents.
Answer quality was assessed by an LLM judge on a 1--5 rubric evaluating factual accuracy and completeness relative to the gold answer (2\,100 total judgments: 300 queries $\times$ 7 system configurations---manifest, three flat-RAG $k$ values, and three metadata-filtered-RAG $k$ values).

\subsection{Retrieval performance}

Table~\ref{tab:overall_eval} summarizes the main comparison across three systems.
Manifest-guided retrieval achieved the highest precision (0.779, 2.2$\times$ over RAG-AF and 3.4$\times$ over flat RAG) and the highest answer-quality score (4.45).
On recall, manifest and RAG-AF were effectively tied overall (0.877 versus 0.876), but this aggregate masks a critical Tier-3 divergence analyzed below.
Notably, RAG-AF achieved more perfect-recall queries than manifest (224/300 versus 214/300), reflecting its higher recall on Tier-1 and Tier-2; however, this came at substantially lower precision (0.352 versus 0.779) and higher document load (7.8 versus 5.8).
Part of the precision gap is structural: both RAG variants always retrieve a fixed number of documents regardless of query complexity, whereas the manifest system adapts its retrieval set (1 document for Tier-1, ${\sim}3.5$ for Tier-2, ${\sim}13$ for Tier-3).

\begin{table}[htbp]
\centering
\small
\caption{Overall comparison between manifest-guided retrieval, metadata-filtered RAG (RAG-AF), and flat RAG at $k{=}10$. Higher is better for score, precision, recall, and perfect recall; lower is better for retrieved documents.}
\label{tab:overall_eval}
\begin{tabular}{lccccc}
\toprule
System & Score & Precision & Recall & Perfect Recall & Docs Retrieved \\
\midrule
Manifest & \textbf{4.45} & \textbf{0.779} & 0.877 & 214/300 & \textbf{5.8} \\
RAG-AF ($k{=}10$) & 4.32 & 0.352 & 0.876 & \textbf{224/300} & 7.8 \\
RAG ($k{=}10$) & 3.85 & 0.231 & 0.647 & 124/300 & 9.7 \\
\bottomrule
\end{tabular}
\end{table}

Table~\ref{tab:tier_eval} breaks down performance by query tier, revealing three key patterns.

First, on Tier-1 and Tier-2 queries, metadata filtering closes the recall gap.
RAG-AF reaches 0.960 and 0.967 recall on Tier-1 and Tier-2 respectively, exceeding manifest (0.870 and 0.913).
This confirms that simple admission-level pre-filtering largely resolves the disambiguation problem for queries that explicitly reference an admission.
However, manifest precision remains 5$\times$ higher on Tier-1 (0.860 versus 0.167) and 1.9$\times$ on Tier-2 (0.840 versus 0.441) because it retrieves only the documents needed.

Second, the clearest manifest advantage appears on Tier-3 longitudinal queries.
When queries span multiple admissions without naming them, admission filtering barely helps (RAG 0.676 $\to$ RAG-AF 0.701).
Manifest recall (0.846) is 21\% higher than RAG-AF, confirming that LLM-guided episode selection is essential for longitudinal queries.

Third, 13 of the 100 Tier-1 queries asked for the chief complaint (gold: \texttt{discharge\_summary.md}); the navigator selected \texttt{triage.md} instead---a plausible but incorrect choice since both documents contain chief-complaint information.
This is a navigator precision limitation rather than a structural failure and is discussed further in Section~\ref{sec:discussion}.

\begin{table}[htbp]
\centering
\small
\caption{Per-tier comparison across manifest-guided retrieval, metadata-filtered RAG (RAG-AF), and flat RAG at $k{=}10$. Manifest advantages are most pronounced on Tier-3 longitudinal queries.}
\label{tab:tier_eval}
\begin{tabular}{clccccc}
\toprule
Tier & System & Score & Precision & Recall & Perfect Recall & Docs \\
\midrule
\multirow{3}{*}{1} & Manifest & 4.76 & \textbf{0.860} & 0.870 & 87/100 & \textbf{1.0} \\
 & RAG-AF ($k{=}10$) & 4.72 & 0.167 & \textbf{0.960} & \textbf{96/100} & 6.6 \\
 & RAG ($k{=}10$) & 3.88 & 0.069 & 0.640 & 64/100 & 9.7 \\
\midrule
\multirow{3}{*}{2} & Manifest & \textbf{4.47} & \textbf{0.840} & 0.913 & 74/100 & \textbf{3.5} \\
 & RAG-AF ($k{=}10$) & 4.28 & 0.441 & \textbf{0.967} & \textbf{91/100} & 7.0 \\
 & RAG ($k{=}10$) & 3.88 & 0.193 & 0.626 & 30/100 & 9.7 \\
\midrule
\multirow{3}{*}{3} & Manifest & \textbf{4.11} & \textbf{0.638} & \textbf{0.846} & \textbf{53/100} & 13.0 \\
 & RAG-AF ($k{=}10$) & 3.95 & 0.449 & 0.701 & 37/100 & 9.6 \\
 & RAG ($k{=}10$) & 3.80 & 0.431 & 0.676 & 30/100 & 9.7 \\
\bottomrule
\end{tabular}
\end{table}

\subsection{Sensitivity and ablation}

Table~\ref{tab:rag_k} shows $k$-sensitivity for both RAG variants.
Even at $k{=}15$, RAG-AF surpasses manifest recall (0.912 versus 0.877)---but at the cost of nearly double the documents (9.8 versus 5.8) and less than half the precision (0.305 versus 0.779).
Manifest matches RAG-AF at $k{=}10$ on recall (0.877 versus 0.876) with 2.2$\times$ higher precision and fewer documents.
Increasing $k$ further yields diminishing returns for RAG-AF while degrading precision.

\begin{table}[htbp]
\centering
\small
\caption{RAG sensitivity to $k$ for both flat and metadata-filtered (RAG-AF) variants. Manifest results (from Table~\ref{tab:overall_eval}) are shown for reference. Score denotes a supporting LLM-judge answer-quality rating (1--5 scale).}
\label{tab:rag_k}
\begin{tabular}{lccccc}
\toprule
System & Precision & Recall & Perfect Recall & Docs Retrieved & Score \\
\midrule
RAG ($k{=}5$)  & 0.291 & 0.439 & 59/300 & 5.0 & 3.28 \\
RAG ($k{=}10$) & 0.231 & 0.647 & 124/300 & 9.7 & 3.85 \\
RAG ($k{=}15$) & 0.186 & 0.738 & 165/300 & 14.3 & 3.95 \\
\midrule
RAG-AF ($k{=}5$)  & 0.466 & 0.768 & 179/300 & 4.6 & 4.11 \\
RAG-AF ($k{=}10$) & 0.352 & 0.876 & 224/300 & 7.8 & 4.32 \\
RAG-AF ($k{=}15$) & 0.305 & 0.912 & 244/300 & 9.8 & 4.33 \\
\midrule
Manifest       & \textbf{0.779} & 0.877 & 214/300 & \textbf{5.8} & \textbf{4.45} \\
\bottomrule
\end{tabular}
\end{table}

To characterize the contribution of each structural layer, we constructed a four-level ablation spectrum (Table~\ref{tab:ablation_spectrum}) ranging from pure dense retrieval to full manifest-guided navigation.

Table~\ref{tab:ablation_spectrum} reveals the contribution of each structural layer.
Adding explicit admission metadata (RAG $\to$ RAG-AF) is the single largest recall improvement ($+$0.229), but it only works when the query text contains admission identifiers (all 100 Tier-1, all 100 Tier-2, but only 21 of 100 Tier-3 queries).
LLM-based episode selection (RAG-AF $\to$ Manifest L1) pushes recall to 0.977 by handling Tier-3 queries that lack explicit metadata, but precision collapses to 0.238 because all documents within selected admissions are loaded.
Document-level filtering (Manifest L1 $\to$ L1+L2) recovers precision to 0.779---a 3.3$\times$ improvement---with a modest recall trade-off ($-$0.100).
This confirms that the document-level manifest is the primary source of the precision advantage.

\begin{table}[htbp]
\centering
\small
\caption{Structural awareness ablation. Each row uses progressively richer structural information from the clinical record.}
\label{tab:ablation_spectrum}
\begin{tabular}{llcccc}
\toprule
System & Structural awareness & Precision & Recall & Perfect Recall & Docs \\
\midrule
RAG ($k{=}10$) & Dense retrieval only & 0.231 & 0.647 & 124/300 & 9.7 \\
RAG-AF ($k{=}10$) & + admission metadata & 0.352 & 0.876 & 224/300 & 7.8 \\
Manifest L1 only & + LLM episode selection & 0.238 & \textbf{0.977} & \textbf{276/300} & 22.2 \\
Manifest L1+L2 & + LLM document selection & \textbf{0.779} & 0.877 & 214/300 & \textbf{5.8} \\
\bottomrule
\end{tabular}
\end{table}

\subsection{Qualitative example}

The manifest-based approach also produced directly interpretable retrieval traces.
For query Q289, a Tier-3 longitudinal query asking how a patient's lactate levels changed from an initial craniotomy admission to a later readmission for surgical site infection, the system resolved the patient from the query text, selected the two relevant admissions, and retrieved exactly two documents: \texttt{lab\_blood\_gas.md} from each admission.
This trace achieved both precision and recall of 1.0.
The resulting answer cited specific lactate values (2.2--3.0\,mmol/L during the craniotomy admission versus 0.9\,mmol/L at readmission), demonstrating that the manifest hierarchy enables clinically precise longitudinal reasoning with minimal document retrieval.

\begin{table}[htbp]
\centering
\small
\caption{Example manifest trace for Q289. The trace shows the selected patient, admission nodes, and leaf documents used to answer a longitudinal lactate-monitoring query on real MIMIC-IV data.}
\label{tab:trace_example}
\begin{tabular}{p{0.22\linewidth}p{0.68\linewidth}}
\toprule
Field & Value \\
\midrule
Query & How did S19768128's lactate levels change from the initial craniotomy admission to the later readmission for surgical site infection? \\
Selected patient & \texttt{S19768128} (resolved from query text) \\
Selected admissions & \texttt{hadm\_20298053}, \texttt{hadm\_20946200} \\
Selected documents & \texttt{lab\_blood\_gas} (from both admissions) \\
Retrieved files & \texttt{S19768128/hadm\_20298053/lab\_blood\_gas.md} \newline \texttt{S19768128/hadm\_20946200/lab\_blood\_gas.md} \\
Trace quality & Precision = 1.0; Recall = 1.0 \\
\bottomrule
\end{tabular}
\end{table}

\section{Discussion}
\label{sec:discussion}
This work makes four conceptual contributions to the design of AI systems for healthcare. We complement these architectural claims with an empirical validation of the page-indexed memory component on de-identified clinical records from MIMIC-IV. Because the benchmark evaluates document retrieval on a single institution's records using a single LLM family, it should be interpreted as targeted evidence for the retrieval architecture rather than as end-to-end validation of a full hospital operating system. Accordingly, the evaluation supports stronger claims about retrieval quality, longitudinal evidence recovery, and trace interpretability than about OS-level safety enforcement or real-world clinical deployment.

First, we reframe the problem of deploying LLM agents in hospitals as an \textit{infrastructure design problem} rather than a model-capability problem. The central argument---that failures of safety, memory, and coordination arise from architectural mismatch rather than insufficient model intelligence---motivates a different design strategy: constrain the environment rather than rely on the model alone. This perspective aligns agent deployment with established principles from operating-systems research, particularly least-privilege execution and isolation as primary safety mechanisms.

Second, we introduce a \textit{document-mutation coordination model} in which all inter-agent communication is mediated through structured file writes, without direct messaging. This design reduces the inter-agent interface to two filesystem primitives, making the coordination protocol verifiable, auditable, and enforceable through OS-native mechanisms. The event-subscription layer built on top of this model supports reactive coordination, including priority-flagged emergency escalation, without requiring agents to maintain persistent communication channels.

Third, we propose \textit{progressive disclosure via manifest-guided navigation} as an alternative to embedding-based memory retrieval. By representing the document hierarchy through human-readable manifest files and delegating navigation decisions to the agent's language-reasoning capability, the architecture eliminates embedding infrastructure while preserving the hierarchical and longitudinal structure of clinical records. Manifest-guided retrieval also yields inherently interpretable access traces: the sequence of manifests read and the children selected at each level constitutes a human-readable record of the information considered before reasoning.
In our MIMIC-IV evaluation, manifest-guided retrieval achieved a 2.2$\times$ precision advantage over the strongest baseline (metadata-filtered RAG) while matching its overall recall (0.877 versus 0.876) and retrieving 26\% fewer documents (5.8 versus 7.8), providing concrete evidence that this navigation strategy scales to real clinical records with hundreds of documents per patient.

Fourth, we identify \textit{ad hoc task composition} as a clinically essential capability that is absent from existing hospital IT systems and show that the proposed architecture can provide it without bespoke engineering for each use case. Because fixed, pre-programmed workflows cannot accommodate the long-tail variability of clinical practice, patients and physicians routinely encounter needs that installed systems cannot address. The agentic model responds by allowing the agent to reason over the medical skill library and compose novel task sequences on demand, augmented by long-term memory that draws on context accumulated across care episodes. This combination of dynamic skill composition and longitudinal memory makes the proposed system qualitatively more adaptive than static workflow-based platforms.

Several limitations of the current paper warrant acknowledgment.
First, manifest descriptions are themselves LLM-generated and are therefore susceptible to inaccuracy, omission, or temporal drift if the manifest update mechanism fails to fire on a mutation event. Although the maintenance policy is designed to be incremental, a high-frequency update environment---such as an intensive care unit generating continuous sensor data---may produce a volume of mutation events that stresses the $O(L)$ LLM-call-per-update budget. Batching strategies and manifest staleness tolerances require further investigation.

Second, the concurrency-control policy is deliberately conservative.
The current design serializes write-producing reasoning cycles at the patient level through short-lived filesystem leases, which is straightforward to implement and audit on Linux but may reduce throughput when many agents contend for the same high-acuity patient record.
In practice, this trade-off is acceptable for the clinical setting we target because the critical safety requirement is to preserve a coherent patient state during each agent's read--reason--write interval, and the duration of that interval is expected to be short relative to the surrounding care workflow.
Nevertheless, future work should evaluate whether finer-grained locking, non-blocking append queues, or transaction-style batching can preserve the same safety guarantees while improving throughput in unusually high-frequency update environments such as intensive care monitoring.

Third, the architecture assumes that patient data originates within the document collection managed by the proposed system. In practice, hospitals operate established electronic health record (EHR) systems such as Epic or Cerner, and clinical data exists in HL7 FHIR-compliant formats that would need to be translated into the document hierarchy. The design of bidirectional synchronization between the proposed document store and existing EHR infrastructure is beyond the scope of this paper.

Fourth, the evaluation uses a single LLM (Kimi K2.5) for all roles---navigation, answer generation, gold-label production, and judging---which may introduce systematic bias. The LLM judge scores reflect quality assessment by the same model family that generated the answers. Cross-model evaluation and human clinical expert review remain necessary for robust validation.

Fifth, the manifest system's admission-level navigation granularity introduces a structural precision--recall trade-off: when a broad query activates an entire admission, sibling documents not needed for the specific query are loaded alongside relevant ones. This over-retrieval was most pronounced on tier-3 longitudinal queries (precision 0.638 versus tier-1 precision 0.860). Finer-grained navigation, such as document-type filtering within admissions, could mitigate this effect.

We tested this directly by adding admission-level metadata filtering to the RAG baseline (RAG-AF). On tier-1 and tier-2 queries where admission identifiers appear in the query text, metadata-filtered RAG closed the recall gap entirely (0.960 and 0.967 versus manifest 0.870 and 0.913) while narrowing the precision gap. However, on tier-3 longitudinal queries---where admission identifiers are typically absent---metadata filtering provided minimal improvement (0.701 versus flat RAG 0.676), and manifest recall (0.846) remained substantially higher. This confirms that the value of hierarchical navigation increases with query complexity: simple metadata filtering largely closes the gap on single-admission lookups, but cross-admission reasoning requires the kind of LLM-guided episode selection that the manifest hierarchy provides.

Our error analysis revealed two systematic patterns that illuminate the strengths and limitations of each approach. First, embedding-based retrieval systematically failed to disambiguate documents sharing vocabulary but differing by temporal or structural context. Of the 46 queries where flat RAG at $k{=}10$ achieved zero recall, 36 were tier-1 queries---typically targeting a specific document from one admission when the patient had many structurally similar documents across admissions. This pattern is consistent with our metadata-filtering experiment: RAG-AF, which pre-filters by admission identifier, reduced tier-1 zero-recall failures from 36 to 4, confirming that the disambiguation failure is fundamentally a metadata problem, not an embedding quality problem. Manifest-guided retrieval largely avoids this failure mode because it resolves the admission before selecting documents within it, though 13 tier-1 queries still failed due to document-type ambiguity within the correct admission.

Second, the manifest system's over-retrieval on tier-3 queries followed a predictable pattern. Across 100 tier-3 queries, the system retrieved 708 non-gold documents (mean 7.08 per query), with three extreme outliers exceeding 60 extra documents each for patients with many admissions. Despite this over-retrieval, manifest tier-3 precision (0.638) still exceeded RAG-AF at $k{=}10$ (0.449).

Third, the 13 tier-1 manifest failures noted above reveal a limitation of LLM-based navigation: when two document types cover similar content, the navigator may select the wrong one. All 13 involved chief-complaint queries where the gold document was \texttt{discharge\_summary.md} but the navigator chose \texttt{triage.md}---both of which contain chief-complaint information. Possible mitigation includes query-aware document ranking within the selected admission.

\section{Conclusion}
We argue that deploying LLM agents in hospitals is primarily an infrastructure problem rather than a model-capability problem, and we propose the \textit{Agentic Operating System for Hospitals} as a concrete response. By reducing each agent's interface to file read and file write, the architecture delegates safety enforcement to the host operating system, replaces vector retrieval with manifest-guided document navigation, and enables ad hoc task composition across a curated medical skill library. Together, these design choices address structural limitations shared by both general-purpose agent frameworks and fixed-function hospital IT systems. In an evaluation on 100 MIMIC-IV patient records with 300 stratified clinical queries (100 per tier), manifest-guided retrieval achieved 2.2$\times$ higher precision than a metadata-filtered RAG baseline (0.779 versus 0.352) while matching its overall recall, and 3.4$\times$ higher precision than flat RAG (0.779 versus 0.231). Even against the metadata-filtered baseline, manifest retrieval achieved 21\% higher recall on longitudinal queries (0.846 versus 0.701), with its clearest advantages observed on multi-admission queries where simple metadata filtering is insufficient. We offer this design as a foundation for clinical AI infrastructure that earns trust not by assuming safe model behavior, but by making unsafe behavior structurally difficult.

\section*{Author Contributions}
W.Y. conceived the study, developed the architecture, and wrote the manuscript.
H.Q. conducted the experiments and implemented the evaluation pipeline.
B.Z. and C.L. contributed clinical expertise and reviewed the clinical workflow and safety considerations.
Z.H. assisted with literature review and manuscript revision.
X.F., R.Y., and J.D. supervised the project and revised the manuscript.
All authors reviewed and approved the final manuscript.

\section*{Competing Interests}
The authors declare no competing interests.

\section*{Data Availability}
The evaluation was conducted on the MIMIC-IV dataset, version 2.2~\cite{johnson2023mimiciv}, with clinical notes from MIMIC-IV-Note~\cite{johnson2023mimicivnote}, available through PhysioNet~\cite{goldberger2000physionet} (\url{https://physionet.org/content/mimiciv/2.2/}) under a credentialed data use agreement. The evaluation pipeline code is available at \url{https://github.com/[repository-placeholder]}. 


\bibliography{ref}

\begin{thebibliography}{10}
\urlstyle{rm}
\expandafter\ifx\csname url\endcsname\relax
  \def\url#1{\texttt{#1}}\fi
\expandafter\ifx\csname urlprefix\endcsname\relax\def\urlprefix{URL }\fi
\expandafter\ifx\csname doiprefix\endcsname\relax\def\doiprefix{DOI: }\fi
\providecommand{\bibinfo}[2]{#2}
\providecommand{\eprint}[2][]{\url{#2}}

\bibitem{yehudai2025evaluation}
\bibinfo{author}{Mohammadi, M.}, \bibinfo{author}{Li, Y.}, \bibinfo{author}{Lo,
  J.} \& \bibinfo{author}{Yip, W.}
\newblock \bibinfo{title}{Evaluation and benchmarking of {LLM} agents: A
  survey}.
\newblock In \emph{\bibinfo{booktitle}{KDD'25, Proceedings of the 31th ACM
  SIGKDD Conference on Knowledge Discovery and Data Mining}},
  \bibinfo{pages}{6129--6139} (\bibinfo{year}{2025}).

\bibitem{wang2025medicalagents}
\bibinfo{author}{Wang, W.} \emph{et~al.}
\newblock \bibinfo{title}{A survey of {LLM}-based agents in medicine: How far
  are we from baymax?}
\newblock In \emph{\bibinfo{booktitle}{Findings of the Association for
  Computational Linguistics: ACL 2025}}, \bibinfo{pages}{10345–10359}
  (\bibinfo{publisher}{Association for Computational Linguistics},
  \bibinfo{year}{2025}).

\bibitem{yao2023react}
\bibinfo{author}{Yao, S.} \emph{et~al.}
\newblock \bibinfo{title}{React: Synergizing reasoning and acting in language
  models}.
\newblock In \emph{\bibinfo{booktitle}{The Eleventh International Conference on
  Learning Representations (ICLR)}} (\bibinfo{year}{2022}).

\bibitem{shinn2023reflexion}
\bibinfo{author}{Shinn, N.}, \bibinfo{author}{Cassano, F.},
  \bibinfo{author}{Gopinath, A.}, \bibinfo{author}{Narasimhan, K.} \&
  \bibinfo{author}{Yao, S.}
\newblock \bibinfo{journal}{\bibinfo{title}{Reflexion: Language agents with
  verbal reinforcement learning}}.
\newblock {\emph{\JournalTitle{Advances in Neural Information Processing
  Systems 36 (NeurIPS)}}}  (\bibinfo{year}{2023}).

\bibitem{park2023generativeagents}
\bibinfo{author}{Park, J.~S.} \emph{et~al.}
\newblock \bibinfo{title}{Generative agents: Interactive simulacra of human
  behavior}.
\newblock In \emph{\bibinfo{booktitle}{The 36th Annual ACM Symposium on User
  Interface Software and Technology (UIST)}}, \bibinfo{pages}{1--22}
  (\bibinfo{year}{2023}).

\bibitem{openclaw2025}
\bibinfo{author}{Steinberger, P.}
\newblock \bibinfo{title}{{OpenClaw}}.
\newblock \bibinfo{howpublished}{\url{https://github.com/openclaw/openclaw}}
  (\bibinfo{year}{2025}).

\bibitem{lewis2020rag}
\bibinfo{author}{Lewis, P.} \emph{et~al.}
\newblock \bibinfo{title}{Retrieval-augmented generation for
  knowledge-intensive {NLP} tasks}.
\newblock In \emph{\bibinfo{booktitle}{Advances in Neural Information
  Processing Systems 33 (NeurIPS)}}, \bibinfo{pages}{9459--9474}
  (\bibinfo{year}{2020}).

\bibitem{packer2023memgpt}
\bibinfo{author}{Packer, C.} \emph{et~al.}
\newblock \bibinfo{journal}{\bibinfo{title}{{MemGPT}: Towards {LLMs} as
  operating systems}}.
\newblock {\emph{\JournalTitle{arXiv preprint arXiv:2310.08560}}}
  (\bibinfo{year}{2023}).

\bibitem{edge2024graphrag}
\bibinfo{author}{Edge, D.} \emph{et~al.}
\newblock \bibinfo{journal}{\bibinfo{title}{From local to global: A graph {RAG}
  approach to query-focused summarization}}.
\newblock {\emph{\JournalTitle{arXiv preprint arXiv:2404.16130}}}
  (\bibinfo{year}{2024}).

\bibitem{sarthi2024raptor}
\bibinfo{author}{Sarthi, P.} \emph{et~al.}
\newblock \bibinfo{title}{Raptor: Recursive abstractive processing for
  tree-organized retrieval}.
\newblock In \emph{\bibinfo{booktitle}{The Twelfth International Conference on
  Learning Representations (ICLR)}} (\bibinfo{year}{2024}).

\bibitem{gutierrez2024hipporag}
\bibinfo{author}{Gutiérrez, B.~J.}, \bibinfo{author}{Shu, Y.},
  \bibinfo{author}{Gu, Y.}, \bibinfo{author}{Yasunaga, M.} \&
  \bibinfo{author}{Su, Y.}
\newblock \bibinfo{journal}{\bibinfo{title}{{HippoRAG}: Neurobiologically
  inspired long-term memory for large language models}}.
\newblock {\emph{\JournalTitle{Advances in neural information processing
  systems (NeurIPS)}}}  (\bibinfo{year}{2024}).

\bibitem{sendak2020clinicalworkflow}
\bibinfo{author}{Sendak, M.~P.} \emph{et~al.}
\newblock \bibinfo{journal}{\bibinfo{title}{Presenting machine learning model
  information to clinical end users with model facts labels}}.
\newblock {\emph{\JournalTitle{npj Digital Medicine}}}
  \textbf{\bibinfo{volume}{3}}, \bibinfo{pages}{41},
  \doiprefix\url{10.1038/s41746-020-0253-3} (\bibinfo{year}{2020}).

\bibitem{nagendran2020aiworkflow}
\bibinfo{author}{Nagendran, M.} \emph{et~al.}
\newblock \bibinfo{journal}{\bibinfo{title}{Artificial intelligence versus
  clinicians: systematic review of design, reporting standards, and claims of
  deep learning studies}}.
\newblock {\emph{\JournalTitle{BMJ}}} \textbf{\bibinfo{volume}{368}},
  \bibinfo{pages}{m689}, \doiprefix\url{10.1136/bmj.m689}
  (\bibinfo{year}{2020}).

\bibitem{rajpurkar2022healthcareai}
\bibinfo{author}{Rajpurkar, P.}, \bibinfo{author}{Chen, E.},
  \bibinfo{author}{Banerjee, O.} \& \bibinfo{author}{Topol, E.~J.}
\newblock \bibinfo{journal}{\bibinfo{title}{Ai in health and medicine}}.
\newblock {\emph{\JournalTitle{Nature Medicine}}}
  \textbf{\bibinfo{volume}{28}}, \bibinfo{pages}{31--38},
  \doiprefix\url{10.1038/s41591-021-01614-0} (\bibinfo{year}{2022}).

\bibitem{ash2004consequences}
\bibinfo{author}{Ash, J.~S.}, \bibinfo{author}{Berg, M.} \&
  \bibinfo{author}{Coiera, E.}
\newblock \bibinfo{journal}{\bibinfo{title}{Some unintended consequences of
  information technology in health care: the nature of patient care information
  system-related errors}}.
\newblock {\emph{\JournalTitle{Journal of the American Medical Informatics
  Association}}} \textbf{\bibinfo{volume}{11}}, \bibinfo{pages}{104--112},
  \doiprefix\url{10.1197/jamia.M1471} (\bibinfo{year}{2004}).

\bibitem{sutton2020cdss}
\bibinfo{author}{Sutton, R.~T.} \emph{et~al.}
\newblock \bibinfo{journal}{\bibinfo{title}{An overview of clinical decision
  support systems: benefits, risks, and strategies for success}}.
\newblock {\emph{\JournalTitle{npj Digital Medicine}}}
  \textbf{\bibinfo{volume}{3}}, \bibinfo{pages}{17},
  \doiprefix\url{10.1038/s41746-020-0221-y} (\bibinfo{year}{2020}).

\bibitem{panch2019inconvenient}
\bibinfo{author}{Panch, T.}, \bibinfo{author}{Mattie, H.} \&
  \bibinfo{author}{Celi, L.~A.}
\newblock \bibinfo{journal}{\bibinfo{title}{The ``inconvenient truth'' about
  {AI} in healthcare}}.
\newblock {\emph{\JournalTitle{npj Digital Medicine}}}
  \textbf{\bibinfo{volume}{2}}, \bibinfo{pages}{77},
  \doiprefix\url{10.1038/s41746-019-0155-4} (\bibinfo{year}{2019}).

\bibitem{holzinger2022trustworthy}
\bibinfo{author}{Holzinger, A.} \emph{et~al.}
\newblock \bibinfo{journal}{\bibinfo{title}{Information fusion as an
  integrative cross-cutting enabler to achieve robust, explainable, and
  trustworthy medical artificial intelligence}}.
\newblock {\emph{\JournalTitle{Information Fusion}}}
  \textbf{\bibinfo{volume}{79}}, \bibinfo{pages}{263--278},
  \doiprefix\url{10.1016/j.inffus.2021.10.007} (\bibinfo{year}{2022}).

\bibitem{johnson2023mimiciv}
\bibinfo{author}{Johnson, A. E.~W.} \emph{et~al.}
\newblock \bibinfo{journal}{\bibinfo{title}{Mimic-iv, a freely accessible
  electronic health record dataset}}.
\newblock {\emph{\JournalTitle{Scientific Data}}}
  \textbf{\bibinfo{volume}{10}}, \bibinfo{pages}{1},
  \doiprefix\url{10.1038/s41597-022-01899-x} (\bibinfo{year}{2023}).

\bibitem{johnson2023mimicivnote}
\bibinfo{author}{Johnson, A. E.~W.}, \bibinfo{author}{Pollard, T.~J.}
  \emph{et~al.}
\newblock \bibinfo{title}{Mimic-iv-note: Deidentified free-text clinical
  notes}.
\newblock
  \bibinfo{howpublished}{\url{https://physionet.org/content/mimic-iv-note/}}
  (\bibinfo{year}{2023}).
\newblock \bibinfo{note}{PhysioNet}.

\bibitem{goldberger2000physionet}
\bibinfo{author}{Goldberger, A.~L.} \emph{et~al.}
\newblock \bibinfo{journal}{\bibinfo{title}{Physiobank, physiotoolkit, and
  physionet: Components of a new research resource for complex physiologic
  signals}}.
\newblock {\emph{\JournalTitle{Circulation}}} \textbf{\bibinfo{volume}{101}},
  \bibinfo{pages}{e215--e220}, \doiprefix\url{10.1161/01.CIR.101.23.e215}
  (\bibinfo{year}{2000}).

\end{thebibliography}

\end{document}